\documentclass[journal]{IEEEtai}
\usepackage[figuresright]{rotating}
\usepackage[linesnumbered,ruled,vlined]{algorithm2e}
\usepackage{algorithmic}
\usepackage{graphicx}
\usepackage{url}
\usepackage{amsmath}
\usepackage{subfigure}
\usepackage{threeparttable}
\usepackage{bm}
\usepackage{float}
\usepackage{color}
\usepackage{booktabs}  
\usepackage{multirow}
\usepackage{setspace}
\usepackage{caption}
\usepackage{epstopdf}
\usepackage{booktabs}
\usepackage{threeparttable}
\usepackage{pifont}
\usepackage{makecell}
\usepackage{cite}
\usepackage{amsfonts}
\usepackage{longtable}
\usepackage{amsmath}

\begin{document}
\title{Maglev-Pentabot: Magnetic Levitation System for Non-Contact Manipulation using Deep Reinforcement Learning}
\author{Guoming Huang$^{*}$, Qingyi Zhou$^{*,\dag}$, Dianjing Liu, Shuai Zhang, Ming Zhou, Zongfu Yu
	\thanks{* Qingyi Zhou and Guoming Huang contributed equally to this work.}
	\thanks{$^{\dag}$ Corresponding author: Qingyi Zhou, qzhou75@wisc.edu.}
	\thanks{Qingyi Zhou, Guoming Huang,  Shuai Zhang and Zongfu Yu are with the Department of Electrical and Computer Engineering, University of Wisconsin-Madison, Dianjing Liu is with Flexcompute, Ming Zhou is with the Department of Electrical Engineering, Stanford University. E-mail: \{Guoming Huang: huangguoming@gbu.edu.cn, Shuai Zhang: szhang565@wisc.edu, Dianjing Liu: dianjing@flexcompute.com, Ming Zhou: mingzhou@stanford.edu, Zongfu Yu: zyu54@wisc.edu\}.}
	}

\markboth{Journal of \LaTeX\ Class Files,~Vol.~18, No.~9, September~2020}%
{How to Use the IEEEtran \LaTeX \ Templates}

\maketitle

\begin{abstract}
Non-contact manipulation has emerged as a transformative approach across various industrial fields. However, current flexible 2D and 3D non-contact manipulation techniques are often limited to microscopic scales, typically controlling objects in the milligram range. In this paper, we present a magnetic levitation system, termed Maglev-Pentabot, designed to address this limitation. The Maglev-Pentabot leverages deep reinforcement learning (DRL) to develop complex control strategies for manipulating objects in the gram range. Specifically, we propose an electromagnet arrangement optimized through numerical analysis to maximize controllable space. Additionally, an action remapping method is introduced to address sample sparsity issues caused by the strong nonlinearity in magnetic field intensity, hence allowing the DRL controller to converge. Experimental results demonstrate flexible manipulation capabilities, and notably, our system can generalize to transport tasks it has not been explicitly trained for. Furthermore, our approach can be scaled to manipulate heavier objects using larger electromagnets, offering a reference framework for industrial-scale robotic applications.
\end{abstract}

\begin{IEEEkeywords}
Non-contact manipulation, magnetic levitation control, industrial robot, deep reinforcement learning, neural network.
\end{IEEEkeywords}

\section{Introduction}
\IEEEPARstart{N}{on-contact} manipulation technology has demonstrated immense potential in industrial and academic applications, particularly in scenarios demanding flexible operations such as smart manufacturing, automated production, semiconductor processing, and medical procedures\cite{abbasi2024autonomous, zhao2021magnetic}. So far, 2D and 3D flexible levitation have been widely studied in microscopic scale, while the controlling objects mostly weigh in milligram range\cite{abbasi2024autonomous}. Hence, 2D and 3D flexible levitation still remains challenging on a macroscopic scale.

Magnetic levitation (maglev) stands out as one of the most promising approaches in the realm of macroscopic scale non-contact manipulation, owing to its strong torque and high flexibility\cite{zhao2021magnetic}. However, due to the nonlinear nature and rapid decaying of magnetic fields, compounded by the difficulties in accurately modeling multi-electromagnet systems\cite{nishino20123d}, these complexities make it challenging to design effective controllers, thereby limiting the widespread adoption of 2D or 3D maglev control. Consequently, overcoming the technical bottlenecks in maglev control would significantly boost the development of industrial automation.

\begin{figure}[htp]
	\centering
	\includegraphics[width=3in]{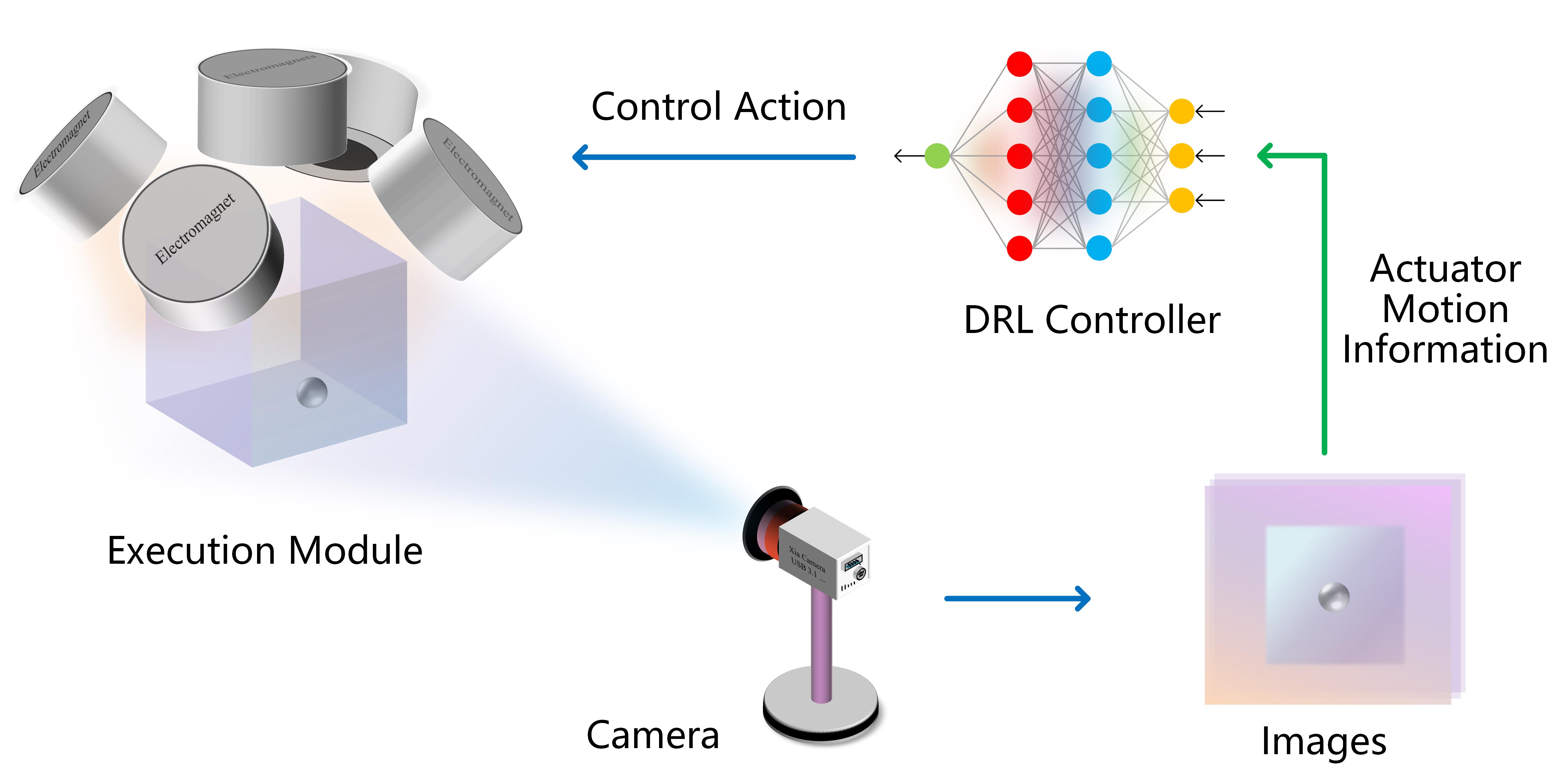}
	\caption{Diagram of Maglev-Pentabot. The Maglev-Pentabot leverages a Deep Reinforcement Learning (DRL) controller as its brain, camera as its vision, and electromagnets coupled with a magnetized ball as the execution module. First, actuator motion information is obtained through the camera. Then, the DRL controller infers control action by inputting this information, thereby achieving control of the actuator.}
	\label{fig1}
\end{figure}

To this end, we develop the Maglev-Pentabot, a maglev-based non-contact manipulation system, which is able to achieve flexible manipulation on a macroscopic scale (see Fig.~\ref{fig1}). The contributions are summarized as follows:
\begin{itemize}
\item To achieve 2D and 3D flexible non-contact manipulation on a  macroscopic scale, we propose a maglev scheme (Maglev-Pentabot) based on DRL. Experimental validation of the prototype demonstrates its potential for accurately and rapidly controlling the actuator (weighting 0.8 g) and transporting heavier objects (weighting 1 g), providing a valuable reference for the development of large-scale, non-contact manipulation robots.
\item Through theoretical analysis, we present electromagnet arrangement method that maximizes the controllable region which is the crucial feature that highlights the application potential for the Maglev-Pentabot.
\item We proposed a comprehensive control scheme based on DRL methods, which, through action remapping and optimized reward function design, enables efficient training of the DRL controller. Additionally, the action remapping method can tackle the challenge of sparse samples due to the highly nonlinear nature of magnetic field intensity.
\end{itemize}

\section{Related work}
\subsection{Non-contact manipulation}
\noindent In fields such as biochemical industry and scientific research, non-contact manipulation technology offers a revolutionary method for moving objects without physical contact\cite{abbasi2024autonomous,nishino20123d,yaseen2017comparative}. Compared to traditional contact-based manipulation, non-contact manipulation exhibits several distinct advantages, such as reduced environmental interference\cite{nishino20123d} and enhanced operational safety\cite{wang2023study,al2022non}. Currently, non-contact manipulation technologies primarily include ultrasonic levitation\cite{seah2014correspondence}, pneumatic levitation\cite{chaos2020robust}, electrostatic levitation\cite{suzuki2010mems}, laser levitation\cite{zheng2020robust}, and magnetic levitation\cite{lin2007intelligent}. Among these, ultrasonic, electrostatic, and laser levitation are limited by their small generated torque, constraining their ability to manipulate heavier objects. Pneumatic levitation, on the other hand, faces difficulties in achieving high-dimensional flexible control due to the complexity of airflow. In contrast, magnetic levitation stands out as the most promising non-contact manipulation method, owing to its higher torque and controllability. It is not only suited for the precise manipulation of small objects but also capable of handling larger, heavier items\cite{wang2023study,kummer2010octomag}. However, the strong nonlinear nature of magnetic forces\cite{boonsatit2016adaptive} presents significant challenges in achieving flexible control of levitated objects on a macroscopic scale. This factor limits the widespread application of magnetic levitation technology in non-contact manipulation\cite{ge2020magnetic,sitti2020pros}.

\subsection{Magnetic levitation control}
\noindent Currently, 1D magnetic levitation control has reached a relatively mature stage, with maglev trains serving as a prime example\cite{yaseen2017comparative}. The vertical control requirements of maglev trains are primarily focused on adaptive control for a single objective (i.e., maintaining a suitable height). Under these conditions, traditional control methods such as PID (Proportional-Integral-Derivative) control or fuzzy control achieve satisfactory results. As such, maglev trains can be regarded as a special case of 1D levitation control systems. In recent years, research into higher dimensional maglev control has been increasing, exploring both 2D\cite{zhu2019flexure,zhang2022modeling,bachovchin2012magnetic,wang2020vertical,marth20132,2010Formulation} and 3D scenarios\cite{abbasi2024autonomous,nishino20123d,1997A,2002Design}. However, these systems generally suffer from slow control response or limited load capacity. The fundamental reason for this lies in the difficulty of accurately modeling the physical systems in high-dimensional maglev setups, coupled with the complexity of the control strategies. Traditional control methods struggle to achieve optimal performance in these scenarios. For instance, PID control often results in significant overshooting, which limits the system's control precision and response speed\cite{nishino20123d}. Robust control, model predictive control, etc. become difficult to implement in this context. Additionally, when the levitation system needs to carry non-magnetic loads, the introduction of such loads significantly alters the system's dynamics, making levitation control even more complex. Therefore, the development of control methods that do not rely on precise physical models becomes particularly important\cite{abbasi2024autonomous}.

\subsection{Deep reinforcement learning}
\noindent Deep Reinforcement Learning (DRL) combines the strengths of deep learning and reinforcement learning, leveraging deep neural networks to process complex high-dimensional data and employing reinforcement learning methods for decision-making. An agent (the entity of DRL method) interacts continuously with its environment, attempting various actions and learning the optimal control strategy from exploratory experiences. In recent years, DRL has demonstrated exceptional performance in fields such as large language models\cite{2020Language}, intelligent Go playing\cite{silver2016mastering}, complex strategy games\cite{mnih2015human}, and robotic control\cite{lobos2018visual}, proving to be a powerful method for solving complex decision-making problems. Among the numerous advanced DRL methods, Proximal Policy Optimization (PPO)\cite{schulman2017proximal} and Soft Actor-Critic (SAC)\cite{haarnoja2018soft} stand out as the most significant. These methods excel in performing complex decision-making tasks by employing strategies such as maximizing action entropy and empirical reuse.

Recently, researchers have begun applying DRL to the challenging field of maglev control. For example, Abssi et al. developed a magnetic microrobot for drug delivery inside human arteries based on DRL\cite{abbasi2024autonomous}, achieving precise in-body control. However, this robot relies on buoyancy for lift, leading to slow transport speeds, limited load capacity, and insufficient flexibility, thus restricting its potential for macroscopic applications. Clearly, DRL stands out as one of the most promising technologies for addressing the challenges of high-dimensional maglev control.

\begin{figure*}
	\centering
	\includegraphics[width=6in]{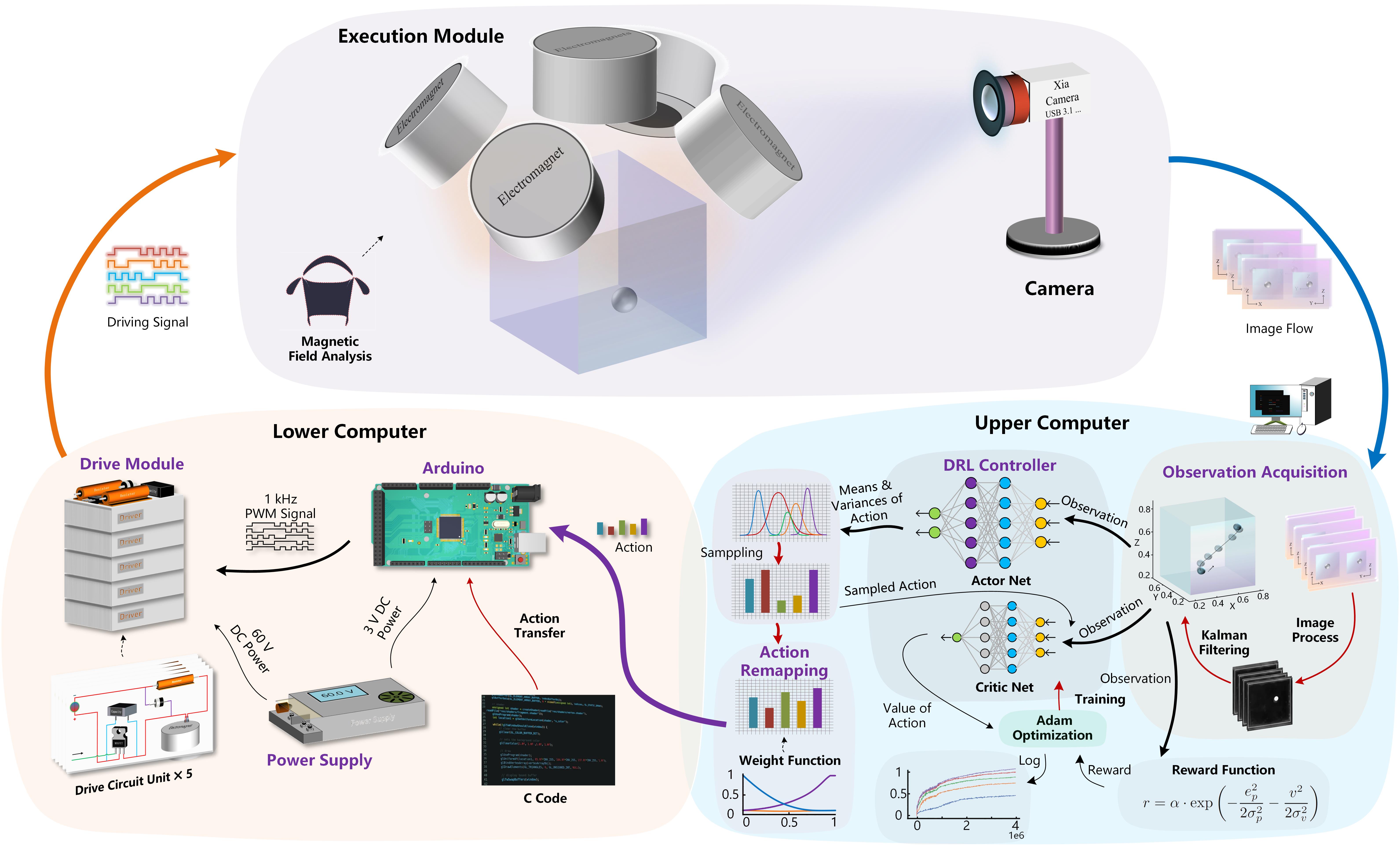}
	\caption{Framework of Maglev-Pentabot. The workflow proceeds as follows: 1. The camera captures images of the actuator's (magnetized ball's) movement and transmits them to the Upper Computer. 2. On the Upper Computer, the motion information of the actuator (such as position, velocity, and acceleration) are extracted through image analysis. The DRL controller then uses this information (as the observation of DRL controller) to infer control action, which are sent to the Arduino on Lower Computer. 3. The Arduino converts the control action into PWM (Pulse Width Modulation)  signal and, via the drive module, converts these PWM signal into drive signal that control the current in the electromagnets (thus controlling the magnetic field strength). This governs the actuator's movement, achieving precise motion control of the magnetized ball.}
	\label{fig2}
\end{figure*}

\section{Maglev-Pentabot}
\noindent In this section, the methodology and implementation of Maglev-Pentabot will be introduced in detail. As shown in Fig.~\ref{fig2}, the overall structure mainly includes three parts: Execution module, Upper Computer and Lower Computer. These three parts collaboratively achieve vision-based feedback for closed-loop maglev control.

\subsection{Execution module}
\noindent The primary challenge in designing the Maglev-Pentabot is how to configure the electromagnet layout to maximize its non-contact manipulation region. To address this, this section analyzes the dynamic stability conditions of a magnetic dipole in a magnetic field, determining the shape of the controllable suspension region under different electromagnet layouts, providing theoretical guidance for the electromagnet arrangement of the Maglev-Pentabot. According to Earnshaw’s theorem\cite{earnshaw1848nature}, stable levitation cannot be achieved in a static magnetic field. Therefore, dynamic equilibrium is a potential approach for achieving stable levitation, but the non-uniformity of the magnetic field makes it impossible to achieve control in all regions. In the following, we first demonstrate that dynamic stability is the fundamental condition for realizing levitation, followed by a simulation of the stabilization regions.

\subsubsection{Basic conditions for stabilized levitation}
\noindent Consider a small magnetic dipole \(\mathbf{m}\) composed of a ferromagnetic material (for example, a permanent magnet). The magnetic field surrounding this dipole is represented as \(\mathbf{B}\), with components \(B_x\), \(B_y\), and \(B_z\). The energy of a small magnetic dipole in a static magnetic field can be calculated as \(U = -\mathbf{m} \cdot \mathbf{B}\). According to \(\nabla^2 U\), it can be demonstrated that a dipole cannot find a stable position in all directions within a static magnetic field. However, if the magnetic field is controllable, the small dipole can be pulled back to a specified position \(P(x_0, y_0, z_0)\) by adjusting the magnetic field whenever it deviates from this position, thus achieving dynamic stability. The specific analysis is as follows.

When the magnetic dipole deviates from its original position \(P(x_0, y_0, z_0)\), the force acting on it consists mainly of two components: the zero-order terms \((\frac{\partial U}{\partial x}, \frac{\partial U}{\partial y}, \frac{\partial U}{\partial z})|_P\), and several first-order terms. Through dynamic control, the change in \(\Delta U\) can induce a force given by \(\frac{\partial U}{\partial x}, \frac{\partial U}{\partial y}, \frac{\partial U}{\partial z}\), which are also first-order terms. However, if the zero-order terms are non-zero, \(\Delta U\) will be overwhelmed. To ensure the effectiveness of the control, dynamic adjustments need to be made around the point where the zero-order terms are equal to zero.

Consider a tiny magnetic dipole \(\mathbf{m}\) composed of ferromagnetic material. Due to the small volume, its moment of inertia can be neglected, and it will always align with the local magnetic field. The dipole \(\mathbf{m}\) is represented as:
\begin{equation}
\mathbf{m} = \frac{k \mathbf{B}}{|\mathbf{B}|}
\end{equation}
where \(k\) denotes the magnitude of the dipole. The energy of the magnetic dipole in the magnetic field can be calculated as:

\begin{equation}
U = -\mathbf{m} \cdot \mathbf{B} = -k \frac{\mathbf{B} \cdot \mathbf{B}}{|\mathbf{B}|} = -k \sqrt{B_x^2 + B_y^2 + B_z^2}
\end{equation}

Now, suppose a disturbance causes the dipole to drift to the point \((x_0 + \Delta x, y_0 + \Delta y, z_0 + \Delta z)\). If the magnetic field does not change with time, then the energy \(U\) will also remain unchanged, and the net force acting on the dipole is given by:

\begin{equation}
	F_x = -\left. \frac{\partial U}{\partial x}\right|_{(x_0 + \Delta x, y_0 + \Delta y, z_0 + \Delta z)}
\end{equation}
\begin{equation}
	F_y = -\left. \frac{\partial U}{\partial y}\right|_{(x_0 + \Delta x, y_0 + \Delta y, z_0 + \Delta z)}
\end{equation}
\begin{equation}
	F_z = -\left. \frac{\partial U}{\partial z}\right|_{(x_0 + \Delta x, y_0 + \Delta y, z_0 + \Delta z)}
\end{equation}

To stabilize the dipole near \(P(x_0, y_0, z_0)\), the magnetic field needs to change, thus producing a new potential energy \(\Delta U\). Compared to \(U\), the potential energy \(\Delta U\) is relatively small:

\begin{equation}
	F_x = -\left. \frac{\partial (U + \Delta U)}{\partial x}\right|_{(x_0 + \Delta x, y_0 + \Delta y, z_0 + \Delta z)}
\end{equation}
\begin{equation}
	F_y = -\left. \frac{\partial (U + \Delta U)}{\partial y}\right|_{(x_0 + \Delta x, y_0 + \Delta y, z_0 + \Delta z)}
\end{equation}
\begin{equation}
	F_z = -\left. \frac{\partial (U + \Delta U)}{\partial z}\right|_{(x_0 + \Delta x, y_0 + \Delta y, z_0 + \Delta z)}
\end{equation}

Performing a Taylor expansion around the point \(P(x_0, y_0, z_0)\) and retaining only the first-order terms, the new net force is:

\begin{equation}
\begin{aligned}
	F_x \approx & -\left. \frac{\partial U}{\partial x}\right|_P - \left. \frac{\partial^2 U}{\partial x^2}\right|_P \Delta x - \left. \frac{\partial^2 U}{\partial x \partial y}\right|_P \Delta y - \\
	& \left. \frac{\partial^2 U}{\partial x \partial z}\right|_P \Delta z - \left. \frac{\partial \Delta U}{\partial x}\right|_P
\end{aligned}
\end{equation}

\begin{equation}
\begin{aligned}
	F_y \approx & -\left. \frac{\partial U}{\partial y}\right|_P - \left. \frac{\partial^2 U}{\partial x \partial y}\right|_P \Delta x - \left. \frac{\partial^2 U}{\partial y^2}\right|_P \Delta y - \\
	& \left. \frac{\partial^2 U}{\partial y \partial z}\right|_P \Delta z - \left. \frac{\partial \Delta U}{\partial y}\right|_P
\end{aligned}
\end{equation}

\begin{equation}
\begin{aligned}
	F_z \approx & -\left. \frac{\partial U}{\partial z}\right|_P - \left. \frac{\partial^2 U}{\partial x \partial z}\right|_P \Delta x - \left. \frac{\partial^2 U}{\partial y \partial z}\right|_P \Delta y - \\
	& \left. \frac{\partial^2 U}{\partial z^2}\right|_P \Delta z - \left. \frac{\partial \Delta U}{\partial z}\right|_P
\end{aligned}
\end{equation}

Without loss of generality, if the zero-order term \(\left. \frac{\partial U}{\partial x}\right|_P \neq 0\), then the new \(F_x\) is bounded, since the other four terms are first-order terms. This means that the dipole cannot stabilize around the point \(P(x_0, y_0, z_0)\). The above derivation proves that \(\left. \frac{\partial U}{\partial x}\right|_P = 0\) is a necessary condition for achieving suspension near \(P(x_0, y_0, z_0)\). Similar conclusions can also be drawn for \(F_y\) and \(F_z\), namely that \(\left. \frac{\partial U}{\partial y}\right|_P = 0\) and \(\left. \frac{\partial U}{\partial z}\right|_P = 0\).

\subsubsection{Analysis of controllable area}
\begin{figure}[htp]
	\centering
	\includegraphics[width=2.8in]{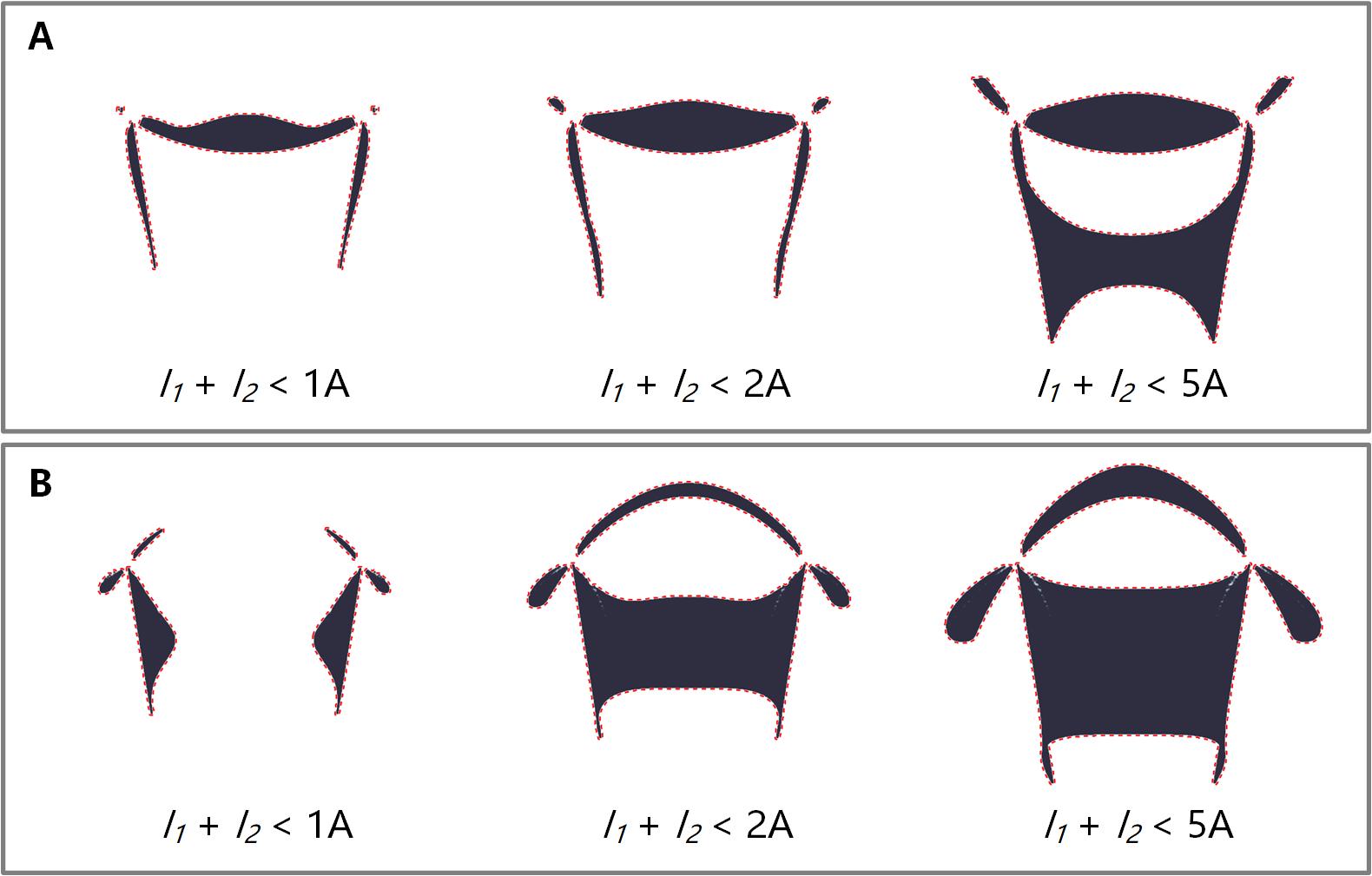}
	\caption{Controllable area in 2D scenario. (A) Magnets with same polarities. (B) Magnets with different polarities. }
	\label{fig3}
\end{figure}

\noindent Based on the above analysis, numerical methods can be employed to determine the controllable regions under different electromagnet configurations. In a 2D scenario, the magnetic field of the Maglev-Pentabot is generated by two electromagnets, with the magnetic dipoles denoted as \( m_1 \) and \( m_2 \). The total magnetic field is calculated by summing the induced magnetic fields. The angle between the axes of the electromagnets and the direction of gravity is set at \( 45^\circ \). This study considers two opposing polarities, using the finite difference method to calculate the force exerted on a magnetized sphere at a given position. By conducting a parameter scan on \( I_1 \) and \( I_2 \), all possible dynamically stable point locations were recorded. The results are shown in Fig.~\ref{fig3}, where different subplots correspond to varying magnetic pole configurations. The distance between the two electromagnets is fixed at 0.15 meters. The objective is to control a magnetic dipole (actuator) with a suspended mass of \( M = 0.8 \, \text{g} \). For other constants, the coefficient is \( k = 1.3 \times 10^{-7} \, \text{J} \cdot \text{m}^3/\text{A} \). A square region surrounding the electromagnets is scanned, with the size of the square domain being \( 0.3 \, \text{m} \times 0.3 \, \text{m} \) and a resolution of 5 mm in both the X and Y directions. From the displayed regions, it can be concluded that a larger current range results in a larger controllable area. Conversely, if the two electromagnets are configured with opposing polarities (arranged in a facing style), the controllable area also increases. Guided by this analysis, reasonable electromagnet configurations were established for both the 2D and 3D scenarios.

\subsubsection{Implementation of actuator}
\begin{figure}[htp]
	\centering
	\includegraphics[width=2.5in]{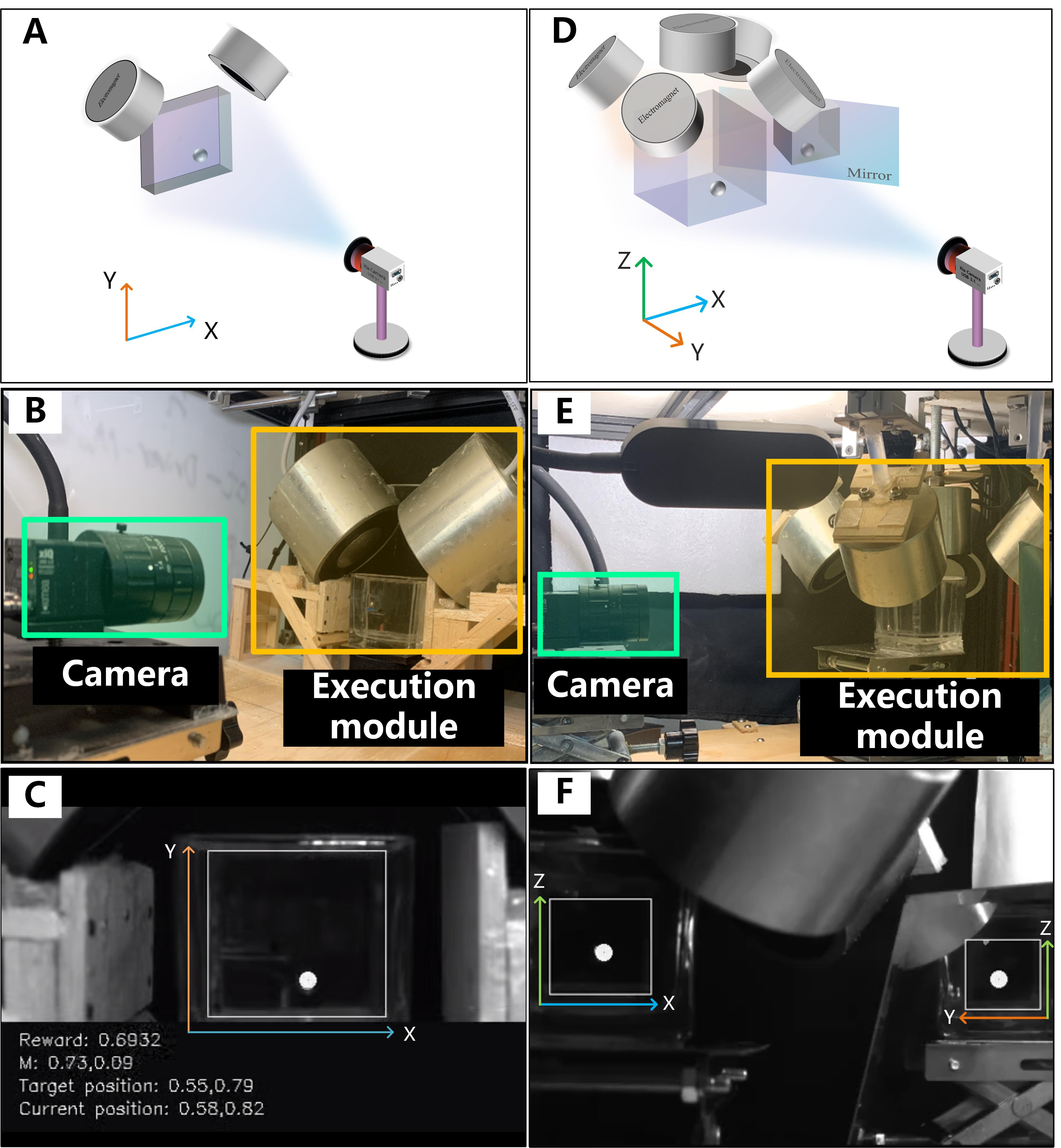}
	\caption{Execution module in 2D and 3D scenarios. (A) Diagram of 2D Execution module. (B) Prototype of 2D Execution module. (C) Camera's field of view in 2D scenario. (D) Diagram of 3D Execution module. (E) Prototype of 3D Execution module. (F) Camera's field of view in 3D scenario.}
	\label{fig4}
\end{figure}

\noindent According to theoretical analysis above, the configuration of the magnets determines the area of the controllable region. To achieve the largest possible controllable area in a 2D scenario, a layout was implemented with two magnets positioned to hang downward (see Fig.~\ref{fig4}A).

These magnets are oriented at a \( 45^\circ \) angle downward relative to each other, creating a significant controllable area beneath them. The magnets operate with different current directions, ensuring that the north and south poles of the actuator are always fixed in orientation. Inspired by the simulation results in the 2D scenario, a configuration using five magnets was adopted for the 3D case, as illustrated in Fig.~\ref{fig4}D. In this setup, four electromagnets are arranged around a central electromagnet, all oriented downward. The central electromagnet is positioned vertically downward, with its current direction opposite to that of the surrounding magnets, which are placed at an approximate angle of \( 60^\circ \) relative to the horizontal. This arrangement forms a configuration akin to a robot with five invisible connecting rods, hence the name Maglev-Pentabot. A camera is positioned 30 cm directly in front of the actuator module, as shown in Fig.~\ref{fig4}B and Fig.~\ref{fig4}E. In the 2D scenario, the Actuator is in the center of the camera’s field of view(see Fig.~\ref{fig4}C). To capture information from all three dimensions, a full reflective mirror is used to provide side-view of the cube to the camera. As depicted in Fig.~\ref{fig4}F, the camera’s field of view allows for both front and side views of the actuator to be captured.

\subsection{DRL controller for High dimensional maglev control}
\noindent Theoretical analysis indicates that dynamic control can be achieved within a specific region. However, due to the complexity of the magnetic field, constructing a high-precision mathematical model for the Maglev-Pentabot, which integrates multiple electromagnets, is challenging. This complexity renders traditional control methods difficult to apply. Hence, this study employs DRL to construct the controller. In the 2D scenario, the Maglev-Pentabot utilizes the PPO method, while in the 3D scenario, the SAC method is employed; Both belong to the Actor-Critic framework\cite{haarnoja2018soft}. Next, we will introduce the design of the DRL controller from three aspects: network structure, observation acquisition, and reward function.


\subsubsection{Network structure}
\begin{figure}[htp]
	\centering
	\includegraphics[width=3.in]{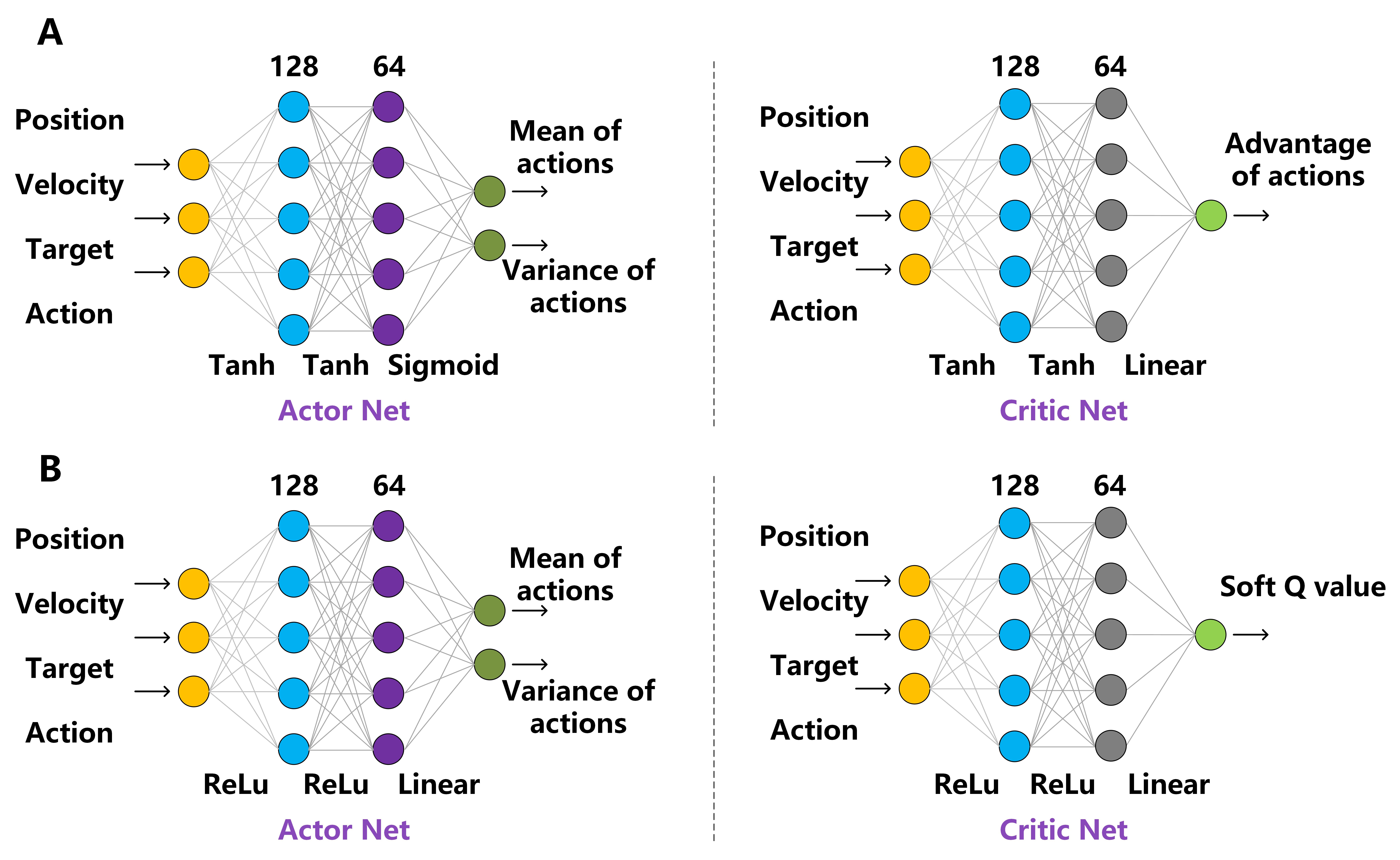}
	\caption{Network structures of DRL agents. (A) The structure of PPO. (B) The structure of SAC.}
	\label{fig5}
\end{figure}

\noindent In the Maglev-Pentabot, the DRL controller is built on the TensorFlow framework, as illustrated in Fig.~\ref{fig5}. The structure of PPO agent is given in Fig.~\ref{fig5}A, while the structure of SAC agent is given in (Fig.~\ref{fig5}B). Both methods consist of fully connected networks for their Actor Net and Critic Net. However, they differ in their choice of activation functions.

\subsubsection{Observation acquisition}
\noindent Intuitively, it may seem better to use images directly as the observation state. However, experiments have shown that, in the case of sparse training samples, employing images requires the network to learn to understand them, which increases the difficulty of network convergence. Therefore, in the design of the state, directly inputting the physical states of the actuator can reduce the learning difficulty of the network. The physical state of the actuator is obtained through image processing methods, which will be detailed in the software design section. The modeling of the observation state is as follows:

\begin{equation}
s_t = \left\{ \begin{bmatrix} p_{t-2} \\ v_{t-2} \\ t_{t-2} \\ a_{t-2} \end{bmatrix},
\begin{bmatrix} p_{t-1} \\ v_{t-1} \\ t_{t-1} \\ a_{t-1} \end{bmatrix},
\begin{bmatrix} p_t \\ v_t \\ t_t \\ a_t \end{bmatrix} \right\}
\end{equation}

where \( t \) denotes the time step, \( s \) represents the observation state, \( p \) indicates the actuator position, \( v \) refers to the actuator velocity, and \( a \) is the action taken by the DRL controller in the previous state. Since the observation state contains information from multiple frames, it implicitly includes the actuator's acceleration information.

\subsubsection{Reward function}
\noindent Theoretically, the reward function can be constructed simply by modeling the distance (position error) between the actuator and the target position. However, preliminary experiments have shown that the strategies trained using this approach are not ideal, especially in terms of maintaining stable levitation after the actuator reaches the target position. To address this issue, this paper introduces a reward function that incorporates velocity constraints and exponential mapping:

\begin{equation}
r = \alpha \cdot \exp\left(-\frac{e_p^2}{2\sigma_p^2} - \frac{v^2}{2\sigma_v^2}\right)
\end{equation}

In this equation, \( \alpha \) represents the reward strength, \( e_p \) and \( v \) denote the position error and ball velocity, respectively, while \( \sigma_p \) and \( \sigma_v \) indicate the reward ranges corresponding to position error and ball velocity. This concept is inspired by the differential term of PID control, utilizing velocity information to guide the policy network in correcting position deviations more quickly and effectively, thereby achieving a more agile levitation effect. Additionally, the purpose of the exponential mapping is to emphasize that the DRL agent can obtain the highest reward when at the target position.

We divided the training process into three phases, with each phase consisting of \( 5 \times 10^6 \) training steps. In the first phase, \( \sigma_p = 0.15 \) and \( \sigma_v = 50 \); in the second phase, \( \sigma_p = 0.1 \) and \( \sigma_v = 30 \); and in the third phase, \( \sigma_p = 0.05 \) and \( \sigma_v = 10 \).

\subsection{Action remapping method}
\noindent The physical property of magnetic field intensity being inversely proportional to the cube of the distance between the magnets and the actuator leads to a rapid increase in magnetic force as the actuator approaches the magnets during the training of the DRL controller. This sharp increase in magnetic force can "trap" the actuator within a small area near the magnets, thereby restricting its range of motion. As the controller interacts with the environment, the exploration space is limited, and the learned policies tend to favor these local regions, resulting in a lack of experience in areas farther from the magnets. This further hinders effective exploration during the training process. The strong nonlinear characteristics of the magnetic field narrow the solution space for the control strategy, making it difficult for the DRL controller to converge.

\begin{figure}[htp]
	\centering
	\includegraphics[width=2.in]{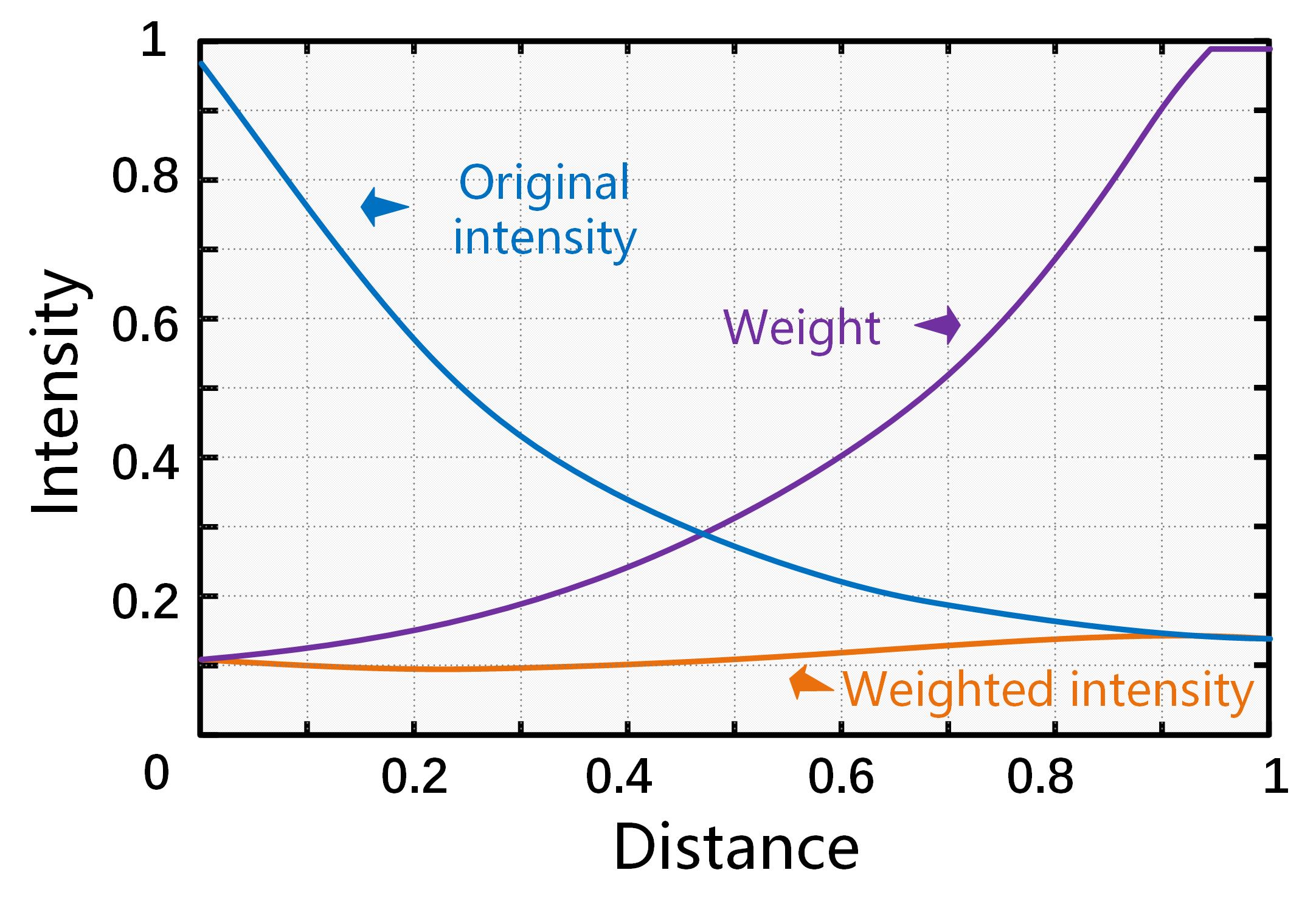}
	\caption{Modification of magnetic field intensity. The original intensity (blue curve) is flattened to the weighted intensity (orange curve) by multiplying it with the distance-based weight (purple curve).}
	\label{fig6}
\end{figure}


To this end, we propose an action remapping method that can reduce the impact of nonlinear characteristics of magnetic field intensity. The core idea is to construct a weight function based on the distance between the actuator and the magnets, gradually modifying the magnetic force that increases sharply with distance. As shown in Fig.~\ref{fig6}, we have constructed a weight mapping curve, which uses distance as the independent variable to smooth the transition of magnetic field intensity growth. The remapping formula is:
\begin{equation}
f_{act} = w \times f_o
\end{equation}
where \(w\) denotes the weight used to map the original magnetic field intensity \(f_o\) to the desired \(f_{act}\). \(w\) is modeled as:
\begin{equation}
w = \min(d_m^2, 1), \quad d \in [0, 1]
\end{equation}
where \(d\) represents the distance between the actuator and the origin. The weight increases rapidly at small distances but remains stable at larger distances, thus avoiding the phenomenon of abrupt increases in magnetic force. The original magnetic field intensity \(f_o\) exhibits an inverse cubic relationship:
\begin{equation}
f_o = \frac{1}{d_m^3}, \quad d_m \in [0, 1]
\end{equation}
where \(d_m\) denotes the distance between the actuator and the magnets.

This remapping effectively mitigates the nonlinear characteristics of the magnetic field intensity, ensuring that the magnetic force does not increase sharply with distance, thus providing the controller with a larger operational space. Experimental results indicate that this design significantly enhances the probability of the small ball reaching other region during training, increasing the exploration capability of the controller. By introducing the weight function, the nonlinearity of the magnetic field is smoothed, reducing the complexity of the control strategy and ultimately overcoming the convergence challenges faced by the DRL controller in high-dimensional magnetic levitation control tasks. Furthermore, the proposed scheme in this section can serve as a  reference for designing DRL-based control systems in other similar scenarios.

\begin{figure}[htp]
	\centering
	\includegraphics[width=3. in]{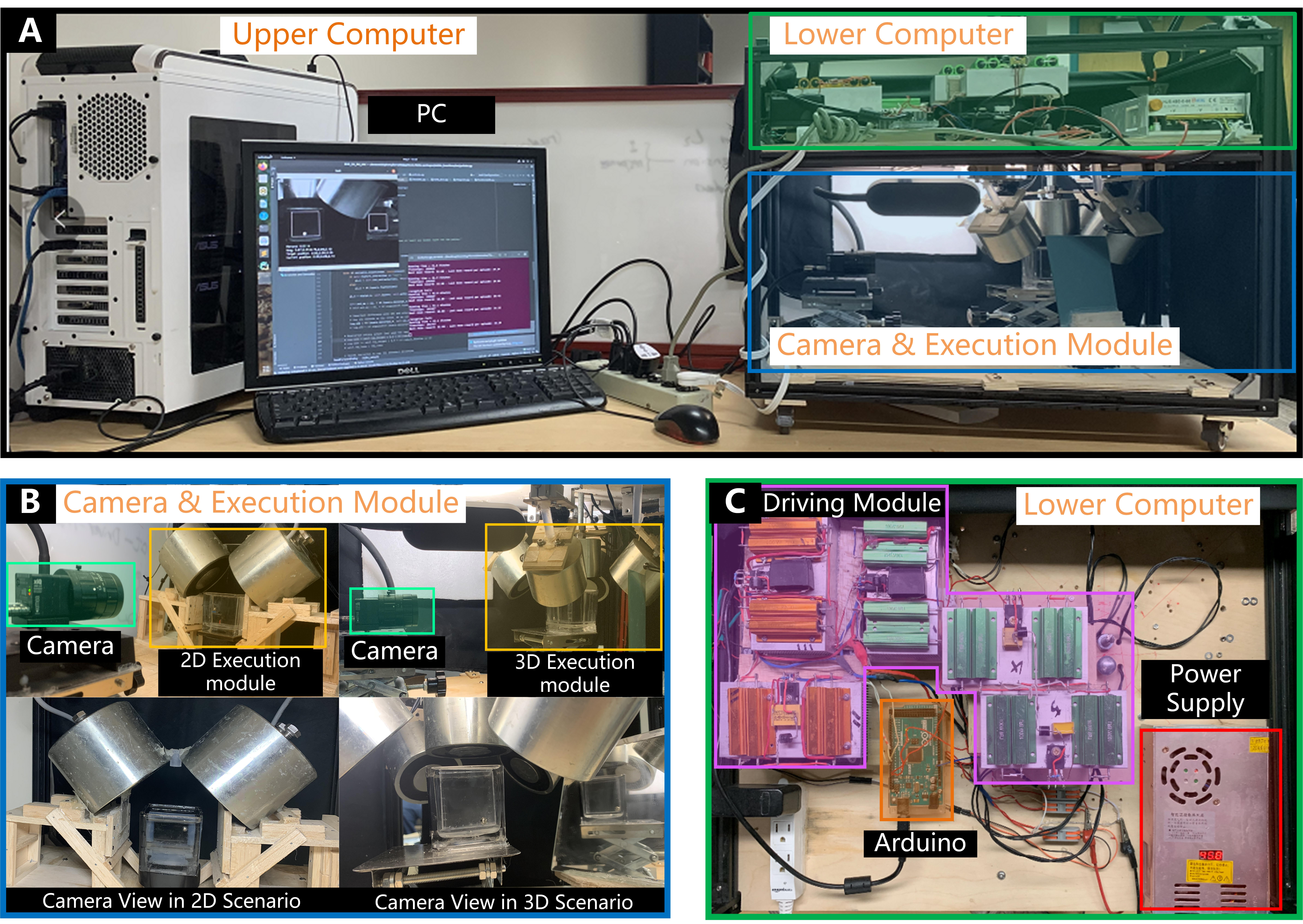}
	\caption{Prototype of Maglev-Pentabot. (A) Overall photo of Maglev-Pentabot. (B) Camera and Execution Module. (C) Lower Computer}
	\label{fig7}
\end{figure}

\section{Experiment}
\noindent The physical prototype of the Maglev-Pentabot is shown in Fig.~\ref{fig7}. This section will validate the performance of the Maglev-Pentabot through three experiments: a multi-target magnetic levitation experiment in a 2D scenario, a transportation experiment, and a multi-target levitation experiment in a 3D scenario. The main configuration of the Maglev-Pentabot includes a PC equipped with an Intel Core i7-13700K CPU, a GTX 1080 GPU, and 16 GB of RAM; an Arduino Mega 2560 with a PWM modulation frequency of 1 kHz; and a control cycle of 10 ms. The drive voltage is set to 60 V.

\subsection{2-D multi-target levitation}
\noindent The purpose of this experiment is to have the Maglev-Pentabot move the actuator to the target position as quickly as possible while maintaining stable levitation, with random switching of the levitation target position. The actuator will remain in place until a new target position appears, at which point it will move to the new location.

\begin{figure}[htp]
	\centering
	\includegraphics[width=3. in]{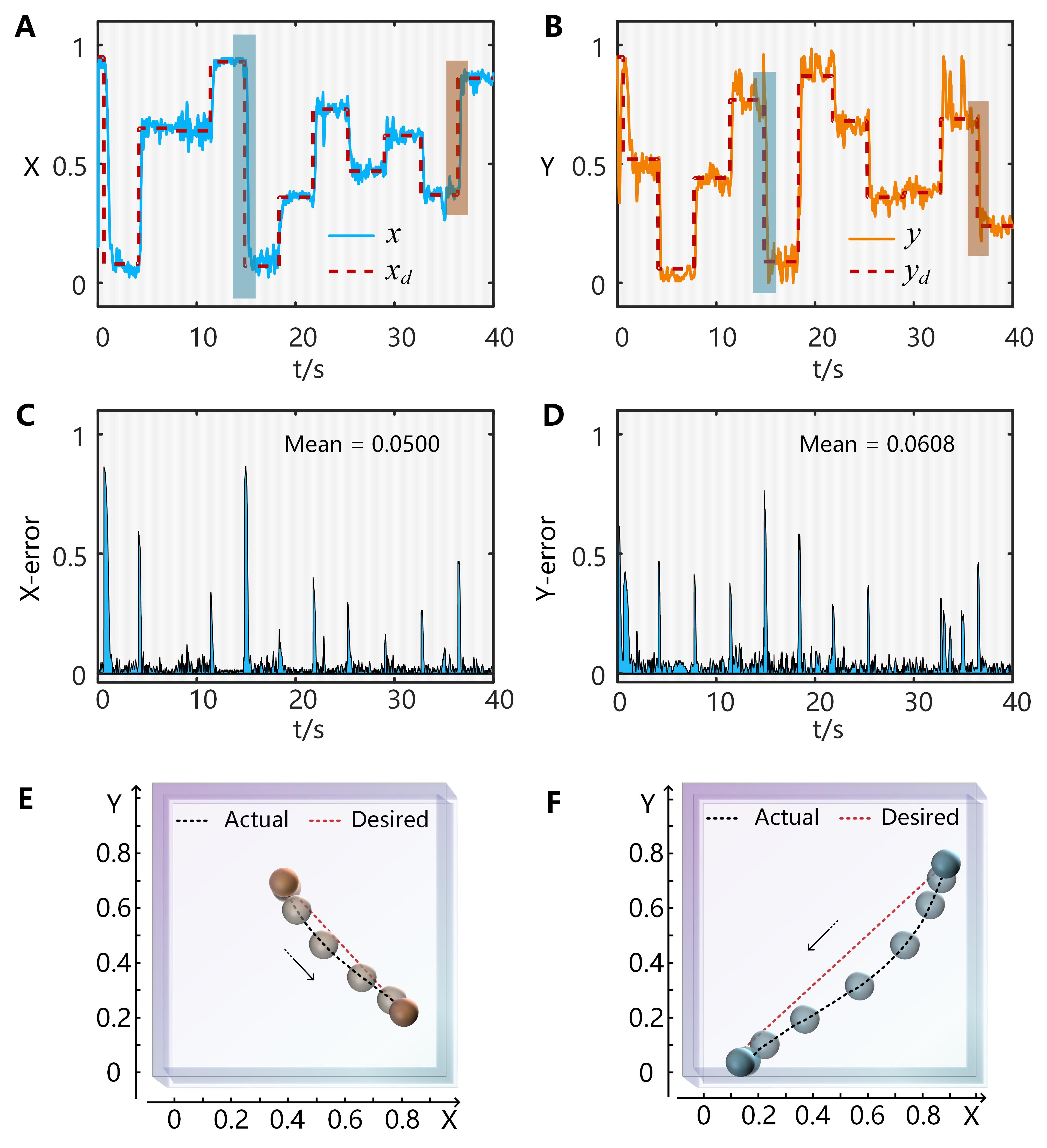}
	\caption{Multi-Target levitation control in 2D scenario. (A) x-direction position of Actuator. (B) y-direction position of Actuator. (C) Control error in x-direction. (D) Control error in y-direction. (E) Actuator trajectory corresponding to the time interval marked by the blue block. (F) Actuator trajectory corresponding to the time interval marked by the orange block.}
	\label{fig8}
\end{figure}

\begin{figure*}
	\centering
	\includegraphics[width=5.2 in]{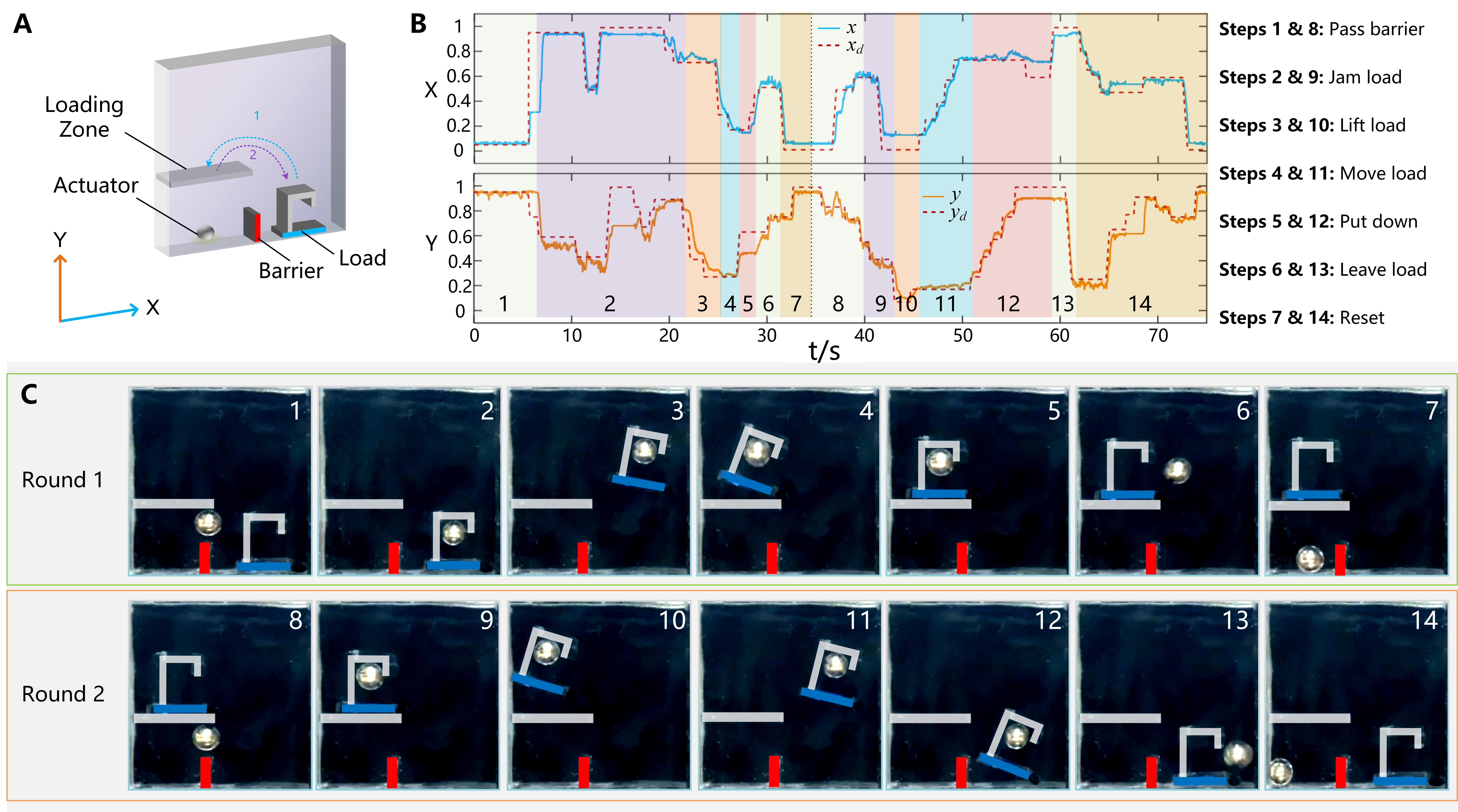}
	\caption{Load handling in 2D scenario. (A) Diagram of handling task. (B) Actuator trajectories of x and y directions, respectively. (C) Images of each step during the manipulation.}
	\label{fig9}
\end{figure*}

The experimental results are illustrated in Fig.~\ref{fig8}, where the actual trajectories of the ball in the x and y directions are plotted as solid lines in Fig.~\ref{fig8}A and Fig.~\ref{fig8}B, respectively, while the error graphs are shown in Fig.~\ref{fig8}C and Fig.~\ref{fig8}D. It is evident that the Maglev-Pentabot achieves high levitation accuracy, with relative mean errors (absolute error values divided by the controllable range) of 0.05 in the X direction and 0.0608 in the Y direction. Since overcoming gravity is the primary challenge for levitation, the average relative error in the Y direction (vertical) is slightly higher. Fig.~\ref{fig8}E and Fig.~\ref{fig8}F present the motion trajectories during two target position switches. The trajectory in Fig.~\ref{fig8}E corresponds to the time interval marked by the blue block in Fig.~\ref{fig8}A and Fig.~\ref{fig8}B, while Fig.~\ref{fig8}F represents the trajectory during the time interval marked by the orange block.

From Fig.~\ref{fig8}A and Fig.~\ref{fig8}B, it can be observed that the time required for the actuator to move from one position to another is very short. The actuator's movement speed reaches 62 mm/s in Fig.~\ref{fig8}E, which is significantly faster than that of other similar magnetic levitation systems \cite{abbasi2024autonomous,nishino20123d}. Additionally, Fig.~\ref{fig8}E and Fig.~\ref{fig8}F demonstrate that the trajectories of the actuator during rapid position switching are not linear but resemble the paths of fastest descent. It is clear that the DRL controller first allows gravity to accelerate the actuator, enabling it to gain speed quickly at the onset. Subsequently, the controller "pulls" the actuator to "whip" it towards the target position, thus enhancing the initial acceleration to increase the overall average speed and achieve the shortest travel time. This demonstrates that the DRL controller can infer an effective control strategy during the training process.

\subsection{2-D transportation}
\noindent Non-contact transportation is a primary potential application of the Maglev-Pentabot. This experiment demonstrates a magnetic levitation transportation scenario from real life. As shown in Fig.~\ref{fig9}A, the actuator serves as a carrier for non-magnetic loads, continuously controlling its position to transport the target device to a specified location. During this process, the levitation target position remains continuous, similar to the operation of a flexible robotic arm's end effector moving along a continuous trajectory.

The experiment consists of two stages. In the first stage, the actuator will first transport the load from the lower region to the upper region and then return to its original position in the lower left corner after successful delivery. The second stage involves transporting the load from the upper region back to the lower region, with the actuator returning to its original position after the delivery is complete.

It is noteworthy that the DRL controller did not attempt the transportation operation during the training process. The load weighs 1 g, exceeding the actuator's weight of 0.8 g. Thus, this experiment serves to validate the generalization performance of the DRL controller, particularly in handling heavier objects. Specifically, when the actuator carries a substantial load, the physical dynamics of the robot are altered, leading the DRL controller to encounter numerous unfamiliar and anomalous observations, potentially impairing its ability to infer the correct actions for the transportation task.

The experimental results are illustrated in Fig.~\ref{fig9}B and Fig.~\ref{fig9}C. Fig.~\ref{fig9}B shows the expected and actual trajectories of the actuator, while Fig.~\ref{fig9}C displays images from different stages of the transportation process. Throughout the two phases of transportation, the control errors remain consistently low, particularly during the load unloading phases between steps 4 to 6 and step 7.

\subsection{3-D multi-target levitation}
\begin{figure}[htp]
	\centering
	\includegraphics[width=3. in]{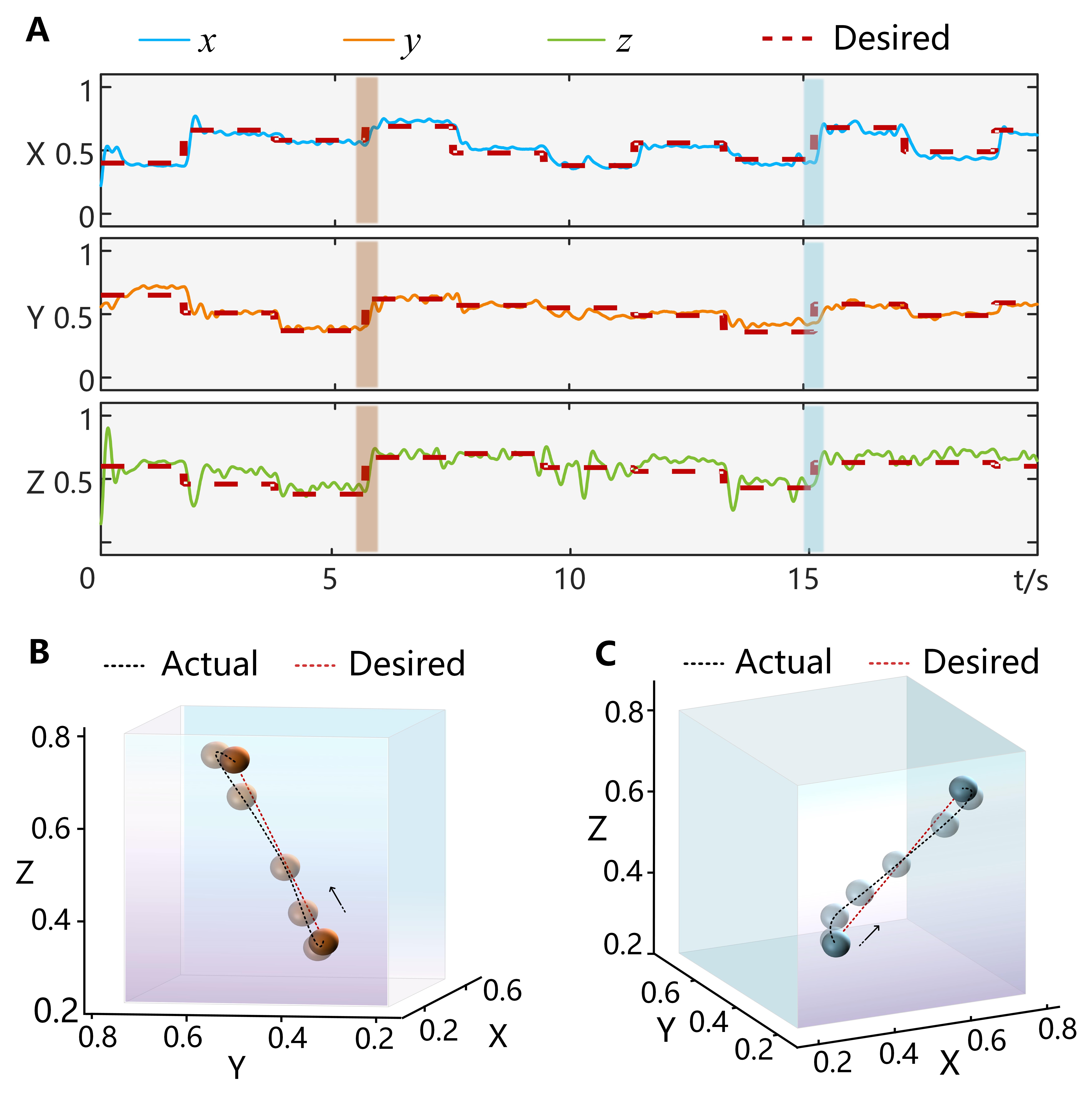}
	\caption{Multi-Target levitation control in 3D scenario. (A) Actuator trajectories of x, y and z directions, respectively. (B) Actuator trajectory corresponding to the time interval marked by the orange block. (C) Actuator trajectory corresponding to the time interval marked by the blue block.}
	\label{fig10}
\end{figure}

\noindent The experimental results are shown in Fig.~\ref{fig10}. Overall, the actuator demonstrates a good ability to track the expected trajectory, with an average moving speed of approximately 96 mm/s, significantly faster than the about 4 mm/s achieved in the 3D magnetic levitation control studied in the literature\cite{abbasi2024autonomous}, while maintaining low error levels.

Furthermore, when comparing the control effects in the x, y, and z directions, it is evident that the control in the horizontal directions is relatively stable, with minimal overshoot, showcasing the robot's excellent levitation control performance. However, the control performance in the z direction is comparatively poorer, as indicated by the fluctuations in the green curve at the 10-second mark. This is primarily due to the need to overcome gravity in the vertical direction, where even slight disturbances in the system (such as unstable Arduino communication) can cause the actuator to jitter.

Fig.~\ref{fig10}B and Fig.~\ref{fig10}C illustrate the actuator's motion trajectories during two target position switches (corresponding to the the time interval marked by the orange and blue blocks in Fig.~\ref{fig10}A). From the actuator's motion trajectory, it can be observed that the controller can also learn complex control strategies to allow the actuator to reach the target point at the fastest speed, rather than taking a straight path.

\section{Discussion}

\subsection{Control performance}
\noindent From the motion trajectory of the multi-target levitation control, it is evident that the actuator's trajectory is similar to the most rapid descending line. Intuitively, while a straight line represents the shortest path, this scenario requires the electromagnet to pull the actuator upward, which counteracts a portion of the gravitational force in the vertical direction. However, the lateral force exerted by the electromagnet cannot be excessively strong at the initial stage; a stronger force implies a greater upward pull, which hinders the actuator from achieving significant acceleration early on.

Overall, despite the strong nonlinear characteristics of magnetic levitation in both 2D and 3D spaces, the Maglev-Pentabot can achieve flexible non-contact manipulation by leveraging the powerful strategic reasoning capabilities of DRL. More importantly, this paper provides experimental evidence that the Maglev-Pentabot can transport non-magnetic objects through obstacles, even without prior training for such tasks, and can handle loads heavier than the actuator itself on a macroscopic scale. This opens up new avenues for non-contact transportation of non-magnetic objects. The versatility of the Maglev-Pentabot presents opportunities for non-contact manipulation applications.

\subsection{Potential application analysis}
\noindent This section will conduct an in-depth analysis of the potential applications of the Maglev-Pentabot by integrating the previously mentioned experimental results with the fundamental principles of magnetic levitation. Specifically, this study will explore the achievable spatial scale and controllable weight that can be managed by the robot when using large electromagnets as the foundation for the actuator modules. This analysis will provide a theoretical basis for the feasibility of the Maglev-Pentabot in various application scenarios.

Assuming that the Maglev-Pentabot is to control the motion of an object composed of paramagnetic (ferromagnetic) materials, the magnetic dipole moment \( \mathbf{m} \) can be expressed as \( \mathbf{m} = kV\mathbf{B} \), where \( k \) is a coefficient dependent on the material. The acceleration induced by the magnetic field is given by:

\begin{equation}
\mathbf{a} = \frac{k V \cdot \nabla (\mathbf{B} \cdot \mathbf{B})}{\rho V} = \frac{k}{\rho} \nabla (\mathbf{B} \cdot \mathbf{B}) \propto \nabla (\mathbf{B} \cdot \mathbf{B})
\end{equation}

where \( \rho \) represents the material's mass density. Since the material remains unchanged, the ratio \( k / \rho \) is a constant. Now consider two systems: for the small system, the magnetic field is provided by the electromagnet, with a dipole moment strength denoted as \( m_0 \); for the larger system, the dipole moment strength of the electromagnet is represented as \( m_0^{\prime} \) (where \( m_0^{\prime} > m_0 \)). For the small system, the induced acceleration around a radius \( r = r_0 \) is given by:

\begin{equation}
| \mathbf{a} (\mathbf{r}_0)| \propto \nabla (\mathbf{B} \cdot \mathbf{B}) \propto m_0^2 \cdot \nabla \left( \frac{1}{r^6} \right)_{\big| r = r_0} \propto \frac{m_0^2}{r_0^7}
\end{equation}

For the larger system, we can similarly calculate the acceleration around \( r = r_0^{\prime} \):

\begin{equation}
|\mathbf{a} (r_0^{\prime} )| \propto \frac{(m_0^{\prime} )^2}{(r_0^{\prime} )^7}
\end{equation}

By scaling proportionally, we can equate \( |\mathbf{a}(r_0)| \) and \( |\mathbf{a}(r_0^{\prime} )| \), leading to:

\begin{equation}
\frac{r_0^{\prime} }{r_0} = \left( \frac{m_0^{\prime} }{m_0} \right)^{2/7}
\end{equation}

When using the large electromagnet represented by \( m_0^{\prime} \), the above scaling relationship can predict the control range of the Maglev-Pentabot \( (r_0^{\prime}) \). The results from the 3D scenario experiments indicate that the existing small robot can control an area of \( 3.5 \, \text{cm}^3 \) with a mass of \( 0.8 \, \text{g} \). Therefore:
\begin{itemize}
\item {When employing a large lifting electromagnet (with a cross-sectional diameter of \( 240 \, \text{cm} \), height of \( 30 \, \text{cm} \), and a maximum operating current of \( 154 \, \text{A} \)) to construct the Delta robot, the robot can levitate and manipulate objects weighing \( 26.2 \, \text{kg} \) within a volume of \( (1.1 \, \text{m})^3 \).}
\item {When using a tokamak coil with a super strong magnetic field (with a cross-sectional area of \( (10.5 \, \text{m})^2 \), height of \( 18.6 \, \text{m} \), and a maximum operating current of \( 75 \, \text{kA} \)), the robot can achieve levitation and manipulation of objects weighing \( 3.8 \times 10^5 \, \text{kg} \) within a volume of \( (27.3 \, \text{m})^3 \).}
\end{itemize}

\subsection{Convergence of PPO controller and SAC controller}
\noindent Through experiments, it was found that the PPO cannot converge in 3D scenarios, whereas the SAC can. PPO is a policy gradient-based deep DRL method, which ensures training stability by restricting the magnitude of policy updates. Its objective function is typically expressed as follows:
\begin{equation}
L(\theta) = \mathbb{E}_{\tau \sim \pi_{\theta_{\text{old}}}} \left[ \min\left(r_t(\theta) A_t,  \text{clip}(r_t(\theta), \epsilon) A_t \right) \right]
\end{equation}

where \(\text{clip}(\theta, \epsilon) = \text{clip}(r_t(\theta), 1 - \epsilon, 1 + \epsilon)\), and \(r_t(\theta) = {\pi_\theta(a_t|s_t)}/{\pi_{\theta_{\text{old}}}(a_t|s_t)}\) represents the ratio of the new policy to the old policy. \(A_t\) is the advantage function, and \(\epsilon\) is a small constant used to limit policy updates. By truncating the ratio \(r_t(\theta)\), PPO prevents excessive updates to the policy, ensuring stability during training. However, this results in the old samples being discarded as the policy is updated, leading to lower data utilization efficiency.

In contrast, SAC focuses on maximizing action entropy, which incorporates a maximum entropy term into the policy optimization to enhance exploration capabilities and data utilization efficiency. The objective function of SAC can be expressed as:
\begin{equation}
L(\theta) = \mathbb{E}_{\tau \sim D} \left[ Q(s_t, a_t) + \alpha H\left(\pi\left(\cdot \middle| s_t\right)\right) \right]
\end{equation}

In this equation, \(Q(s_t, a_t)\) is the action-value function, representing the expected return after executing action \(a_t\) in state \(s_t\). \(H(\pi(\cdot|s_t))\) is the entropy of the policy, which encourages exploration. The parameter \(\alpha\) is a coefficient that balances exploration and exploitation, and \(D\) is the set of data samples drawn from the experience replay buffer. Since SAC is a purely offline training method, it can fully utilize experience replay techniques, allowing historical samples to continuously be used for policy optimization, resulting in higher data utilization efficiency.

\subsection{Hardware limitation}
\noindent According to the experiments, the accuracy in the vertical direction (i.e., the z-direction) is slightly lower than that in the horizontal direction (i.e., the x and y directions). This discrepancy is primarily due to the influence of gravity. Even within a 40 mm high space, the small ball can accelerate to approximately 980 mm/s within 0.1 seconds due to gravitational force. Therefore, the magnetic field and the corresponding output current must be rapidly adjusted to overcome this gravitational pull. Such stringent requirements exceed the capabilities of the hardware system (Arduino and PC).

To circumvent the limitations of the hardware performance, we filled the acrylic box with mineral oil during the experiment, which helped reduce the acceleration demands for manipulating the actuator under extreme conditions. Despite these hardware constraints, our DRL controller still achieved a maximum vertical speed of 96 mm/s, which is about 50 times faster than that reported in the literature\cite{nishino20123d}.

The stability of the Maglev-Pentabot's levitation is also affected by interference from the entire power supply and communication systems. Particularly in 3D scenarios, any issue with the Arduino communication can cause significant shaking of the actuator if even one of the five electromagnetic units experiences a malfunction. This situation could be alleviated in the future through optimization of the lower-level control systems or by implementing redundancy designs.

\section{Conclusion}
\noindent This paper develops the Maglev-Pentabot for flexible non-contact manipulation of objects on a macroscopic scale using DRL. To achieve maximum maglev control in 2D and 3D scenarios, the study begins with a theoretical analysis to determine the optimal arrangement of magnets. Meanwhile, we developed an action remapping method and a novel reward function to address the convergence challenges of the DRL controller. The performance of the Maglev-Pentabot is demonstrated through various experiments, confirming its capability for accurate and rapid levitation control. Furthermore, the transportation experiment indicates that the robot can generalize to transport tasks for which it was not specifically trained. Additionally, based on experimental results, an application potential analysis of the Maglev-Pentabot suggests that under ideal hardware conditions, it can transport non-magnetic objects weighing \(3.8 \times 10^5\) kg over a volume exceeding \((27.3 \, \text{m})^3\). This showcases the significant application prospects of the robot in the field of macroscopic non-contact manipulation.

\subsection{Acknowledgments}
\noindent Funding: DARPA Award No. FA8650-20-1-7028, DARPA NLM program Award HR00111820046.

\bibliographystyle{IEEEtran}
\bibliography{IEEEexample}

\begin{thebibliography}{10}
\providecommand{\url}[1]{#1}
\csname url@samestyle\endcsname
\providecommand{\newblock}{\relax}
\providecommand{\bibinfo}[2]{#2}
\providecommand{\BIBentrySTDinterwordspacing}{\spaceskip=0pt\relax}
\providecommand{\BIBentryALTinterwordstretchfactor}{4}
\providecommand{\BIBentryALTinterwordspacing}{\spaceskip=\fontdimen2\font plus
\BIBentryALTinterwordstretchfactor\fontdimen3\font minus
  \fontdimen4\font\relax}
\providecommand{\BIBforeignlanguage}[2]{{%
\expandafter\ifx\csname l@#1\endcsname\relax
\typeout{** WARNING: IEEEtran.bst: No hyphenation pattern has been}%
\typeout{** loaded for the language `#1'. Using the pattern for}%
\typeout{** the default language instead.}%
\else
\language=\csname l@#1\endcsname
\fi
#2}}
\providecommand{\BIBdecl}{\relax}
\BIBdecl

\bibitem{abbasi2024autonomous}
S.~A. Abbasi, A.~Ahmed, S.~Noh, N.~L. Gharamaleki, S.~Kim, A.~M.~B. Chowdhury,
  J.-y. Kim, S.~Pan{\'e}, B.~J. Nelson, and H.~Choi, ``Autonomous 3d positional
  control of a magnetic microrobot using reinforcement learning,'' \emph{Nature
  Machine Intelligence}, vol.~6, no.~1, pp. 92--105, 2024.

\bibitem{zhao2021magnetic}
P.~Zhao, Y.~Jia, J.~Xie, T.~Wang, C.~Zhang, and J.~Fu, ``Magnetic levitation
  for polymer testing using magnet array,'' \emph{Polymer Testing}, vol. 103,
  p. 107361, 2021.

\bibitem{nishino20123d}
T.~Nishino, Y.~Fujitani, N.~Kato, N.~Tsuda, Y.~Nomura, and H.~Matsui, ``3d
  positional control of magnetic levitation system using adaptive control:
  improvement of positioning control in horizontal plane,'' in
  \emph{Intelligent Robots and Computer Vision XXIX: Algorithms and
  Techniques}, vol. 8301.\hskip 1em plus 0.5em minus 0.4em\relax SPIE, 2012,
  pp. 192--197.

\bibitem{yaseen2017comparative}
M.~H. Yaseen, ``A comparative study of stabilizing control of a planer
  electromagnetic levitation using pid and lqr controllers,'' \emph{Results in
  physics}, vol.~7, pp. 4379--4387, 2017.

\bibitem{wang2023study}
X.~Wang and J.~Huang, ``Study on electromagnetic relationship and dynamic
  characteristics of superconducting electrodynamic maglev train on curved
  track,'' \emph{IEEE Transactions on Intelligent Transportation Systems},
  vol.~24, no.~6, pp. 6146--6164, 2023.

\bibitem{al2022non}
I.~I.~I. Al-Nuaimi, M.~N. Mahyuddin, and N.~K. Bachache, ``A non-contact
  manipulation for robotic applications: A review on acoustic levitation,''
  \emph{IEEE Access}, vol.~10, pp. 120\,823--120\,837, 2022.

\bibitem{seah2014correspondence}
S.~A. Seah, B.~W. Drinkwater, T.~Carter, R.~Malkin, and S.~Subramanian,
  ``Correspondence: Dexterous ultrasonic levitation of millimeter-sized objects
  in air,'' \emph{IEEE transactions on ultrasonics, ferroelectrics, and
  frequency control}, vol.~61, no.~7, pp. 1233--1236, 2014.

\bibitem{chaos2020robust}
D.~Chaos, J.~Chac{\'o}n, E.~Aranda-Escol{\'a}stico, and S.~Dormido, ``Robust
  switched control of an air levitation system with minimum sensing,''
  \emph{ISA transactions}, vol.~96, pp. 327--336, 2020.

\bibitem{suzuki2010mems}
Y.~Suzuki, D.~Miki, M.~Edamoto, and M.~Honzumi, ``A mems electret generator
  with electrostatic levitation for vibration-driven energy-harvesting
  applications,'' \emph{Journal of Micromechanics and Microengineering},
  vol.~20, no.~10, p. 104002, 2010.

\bibitem{zheng2020robust}
Y.~Zheng, L.-M. Zhou, Y.~Dong, C.-W. Qiu, X.-D. Chen, G.-C. Guo, and F.-W. Sun,
  ``Robust optical-levitation-based metrology of nanoparticle’s position and
  mass,'' \emph{Physical review letters}, vol. 124, no.~22, p. 223603, 2020.

\bibitem{lin2007intelligent}
F.-J. Lin, L.-T. Teng, and P.-H. Shieh, ``Intelligent adaptive backstepping
  control system for magnetic levitation apparatus,'' \emph{IEEE transactions
  on magnetics}, vol.~43, no.~5, pp. 2009--2018, 2007.

\bibitem{kummer2010octomag}
M.~P. Kummer, J.~J. Abbott, B.~E. Kratochvil, R.~Borer, A.~Sengul, and B.~J.
  Nelson, ``Octomag: An electromagnetic system for 5-dof wireless
  micromanipulation,'' \emph{IEEE Transactions on Robotics}, vol.~26, no.~6,
  pp. 1006--1017, 2010.

\bibitem{boonsatit2016adaptive}
N.~Boonsatit and C.~Pukdeboon, ``Adaptive fast terminal sliding mode control of
  magnetic levitation system,'' \emph{Journal of Control, Automation and
  Electrical Systems}, vol.~27, pp. 359--367, 2016.

\bibitem{ge2020magnetic}
S.~Ge, A.~Nemiroski, K.~A. Mirica, C.~R. Mace, J.~W. Hennek, A.~A. Kumar, and
  G.~M. Whitesides, ``Magnetic levitation in chemistry, materials science, and
  biochemistry,'' \emph{Angewandte Chemie International Edition}, vol.~59,
  no.~41, pp. 17\,810--17\,855, 2020.

\bibitem{sitti2020pros}
M.~Sitti and D.~S. Wiersma, ``Pros and cons: Magnetic versus optical
  microrobots,'' \emph{Advanced Materials}, vol.~32, no.~20, p. 1906766, 2020.

\bibitem{zhu2019flexure}
H.~Zhu, T.~J. Teo, and C.~K. Pang, ``Flexure-based magnetically levitated
  dual-stage system for high-bandwidth positioning,'' \emph{IEEE Transactions
  on Industrial Informatics}, vol.~15, no.~8, pp. 4665--4675, 2019.

\bibitem{zhang2022modeling}
H.~Zhang, Y.~Lou, L.~Zhou, Z.~Kou, and J.~Mu, ``Modeling and optimization of a
  large-load magnetic levitation gravity compensator,'' \emph{IEEE Transactions
  on Industrial Electronics}, vol.~70, no.~5, pp. 5055--5064, 2022.

\bibitem{bachovchin2012magnetic}
K.~D. Bachovchin, J.~F. Hoburg, and R.~F. Post, ``Magnetic fields and forces in
  permanent magnet levitated bearings,'' \emph{IEEE Transactions on Magnetics},
  vol.~48, no.~7, pp. 2112--2120, 2012.

\bibitem{wang2020vertical}
L.~Wang, Z.~Deng, Y.~Li, and H.~Li, ``Vertical--lateral coupling force relation
  of the high-temperature superconducting magnetic levitation system,''
  \emph{IEEE Transactions on Applied Superconductivity}, vol.~31, no.~1, pp.
  1--6, 2020.

\bibitem{marth20132}
E.~Marth, G.~Jungmayr, and W.~Amrhein, ``A 2-d-based analytical method for
  calculating permanent magnetic ring bearings with arbitrary magnetization and
  its application to optimal bearing design,'' \emph{IEEE transactions on
  magnetics}, vol.~50, no.~5, pp. 1--8, 2013.

\bibitem{2010Formulation}
H.~Miyazaki, T.~Ohji, K.~Amei, and M.~Sakui, ``Formulation of a
  three-dimensional movable magnetic levitation system and its performance in
  tests,'' \emph{Journal of the Magnetics Society of Japan}, vol.~34, no.~3,
  pp. 395--400, 2010.

\bibitem{1997A}
T.~Nakamura and M.~B. Khamesee, ``A prototype mechanism for three-dimensional
  levitated movement of a small magnet,'' \emph{IEEE/ASME Transactions on
  Mechatronics}, vol.~2, no.~1, pp. 41--50, 1997.

\bibitem{2002Design}
M.~B. Khamesee, N.~Kato, Y.~Nomura, and T.~Nakamura, ``Design and control of a
  microrobotic system using magnetic levitation,'' \emph{Mechatronics IEEE/ASME
  Transactions on}, vol.~7, no.~1, pp. 1--14, 2002.

\bibitem{2020Language}
T.~B. Brown, B.~Mann, N.~Ryder, M.~Subbiah, and D.~Amodei, ``Language models
  are few-shot learners,'' 2020.

\bibitem{silver2016mastering}
D.~Silver, A.~Huang, C.~J. Maddison, A.~Guez, L.~Sifre, G.~Van Den~Driessche,
  J.~Schrittwieser, I.~Antonoglou, V.~Panneershelvam, M.~Lanctot \emph{et~al.},
  ``Mastering the game of go with deep neural networks and tree search,''
  \emph{nature}, vol. 529, no. 7587, pp. 484--489, 2016.

\bibitem{mnih2015human}
V.~Mnih, K.~Kavukcuoglu, D.~Silver, A.~A. Rusu, J.~Veness, M.~G. Bellemare,
  A.~Graves, M.~Riedmiller, A.~K. Fidjeland, G.~Ostrovski \emph{et~al.},
  ``Human-level control through deep reinforcement learning,'' \emph{nature},
  vol. 518, no. 7540, pp. 529--533, 2015.

\bibitem{lobos2018visual}
K.~Lobos-Tsunekawa, F.~Leiva, and J.~Ruiz-del Solar, ``Visual navigation for
  biped humanoid robots using deep reinforcement learning,'' \emph{IEEE
  Robotics and Automation Letters}, vol.~3, no.~4, pp. 3247--3254, 2018.

\bibitem{schulman2017proximal}
J.~Schulman, F.~Wolski, P.~Dhariwal, A.~Radford, and O.~Klimov, ``Proximal
  policy optimization algorithms,'' \emph{arXiv preprint arXiv:1707.06347},
  2017.

\bibitem{haarnoja2018soft}
T.~Haarnoja, A.~Zhou, K.~Hartikainen, G.~Tucker, S.~Ha, J.~Tan, V.~Kumar,
  H.~Zhu, A.~Gupta, P.~Abbeel \emph{et~al.}, ``Soft actor-critic algorithms and
  applications,'' \emph{arXiv preprint arXiv:1812.05905}, 2018.

\bibitem{earnshaw1848nature}
S.~Earnshaw, ``On the nature of the molecular forces which regulate the
  constitution of the luminiferous ether,'' \emph{Transactions of the Cambridge
  Philosophical Society}, vol.~7, p.~97, 1848.

\end{thebibliography}

\end{document}